\documentclass[letterpaper]{article} 
\usepackage{aaai2026}  
\usepackage{times}  
\usepackage{helvet}  
\usepackage{courier}  
\usepackage[hyphens]{url}  
\usepackage{graphicx} 
\usepackage{natbib}  
\usepackage{caption} 
\usepackage{amsmath}       
\usepackage{amssymb}       
\usepackage{amsfonts}      
\usepackage{graphicx}      
\usepackage{booktabs}      
\usepackage{algorithm}
\usepackage{algpseudocode} 
\usepackage{enumitem}      
\usepackage{xcolor}        

\usepackage{booktabs}     
\usepackage{multirow}     
\usepackage{amssymb}      
\usepackage{graphicx}     

\usepackage{tikz}
\usepackage{caption}
\usepackage{rotating}

\usepackage{newfloat}
\usepackage{listings}
\DeclareCaptionStyle{ruled}{labelfont=normalfont,labelsep=colon,strut=off} 
\floatstyle{ruled}
\newfloat{listing}{tb}{lst}{}
\floatname{listing}{Listing}

\title{Talk2Image: A Multi-Agent System for Multi-Turn Image Generation and Editing}
\author{
    Shichao Ma, Yunhe Guo, Jiahao Su, Qihe Huang, Zhengyang Zhou, Yang Wang
}
\affiliations{
    University of Science and Technology of China
}

\usepackage{bibentry}

\begin{document}

\maketitle

\begin{abstract}
Text-to-image generation tasks have driven remarkable advances in diverse media applications, yet most focus on single-turn scenarios and struggle with iterative, multi-turn creative tasks. Recent dialogue-based systems attempt to bridge this gap, but their single-agent, sequential paradigm often causes intention drift and incoherent edits.
To address these limitations, we present \textbf{Talk2Image}, a novel multi-agent system for interactive image generation and editing in multi-turn dialogue scenarios. Our approach integrates three key components: intention parsing from dialogue history, task decomposition and collaborative execution across specialized agents, and feedback-driven refinement based on a multi-view evaluation mechanism. Talk2Image enables step-by-step alignment with user intention and consistent image editing. Experiments demonstrate that Talk2Image outperforms existing baselines in controllability, coherence, and user satisfaction across iterative image generation and editing tasks. 

\end{abstract}

\section{Introduction}
Recent years have seen remarkable progress in text-to-image (T2I) generative models, especially diffusion-based approaches~\cite{ho2020denoising, podell2023sdxl}, which produce high-quality and diverse images from concise textual prompts. This has driven widespread adoption in art creation, graphic design, advertising, and other domains. Meanwhile, vibrant open-source communities like HuggingFace~\cite{huggingface2018}, Civitai~\cite{civitai2022}, and OpenArt~\cite{OpenArt2021} have accelerated the sharing of models and workflows, broadening the selection of models available to users.

Despite these advances, most T2I models still lack mechanisms for dynamic interactivity and multi-turn control. Users often find it challenging to iteratively refine complex visual intentions through natural language, limiting the progressive articulation of creative goals~\cite{ma2025dialogdraw}. Recent dialogue-based systems, such as GenArtist~\cite{wang2024genartist}, DialogDraw~\cite{ma2025dialogdraw}, and DialogGen~\cite{huang2024dialoggen}, integrate large language models (LLMs) to close the loop between user intention and image generation. However, they typically adopt a single-agent, sequential paradigm, which limits modularity collaboration and often leads to \textit{intention drift} (misalignment with cumulative user goals) and \textit{incoherent edits} (visual inconsistencies across iterations) in complex tasks.
Multi-agent systems (MAS) offer a promising alternative to address these limitations through task decomposition and collaboration \cite{calegari2021logic, cardoso2021review}. 
With the growing availability of specialized generative models on platforms like Civitai, a natural question arises: \textbf{\textit{Can we leverage these resources to develop a MAS framework supporting sustained, multi-turn image generation and editing?}} To turn this vision into reality, three key challenges must be addressed: 
(1) accurately parsing user intention and transforming it into structured, executable prompts; 
(2) decomposing complex tasks into diverse subtasks (e.g., object addition, style modification) and coordinating their execution via multi-agent collaboration; and 
(3) refining outputs through iterative feedback to ensure semantic alignment and visual quality.

To tackle these challenges, we propose \textbf{Talk2Image}, a novel multi-agent system for multi-turn interactive image generation and editing. The system comprises three core components that address key limitations of existing frameworks: (1) a dynamic intention parsing module that synthesizes structured prompts based on dialogue history, directly mitigating intention drift; (2) a modular agent collaboration mechanism enabling task decomposition and specialized execution, with a Directed Acyclic Graph ensuring consistency and valid execution for coherent edits across iterations; and (3) a multi-view evaluation and refinement loop that maintains step-by-step alignment with user intention.

Our contributions are summarized as follows:


\begin{itemize}
    \item We introduce \textbf{Talk2Image}, the first multi-agent system tailored for image generation and editing in multi-turn dialogues.
    \item We design a three-stage architecture encompassing intention parsing, multi-agent collaboration, and a closed-loop feedback mechanism.
    \item We demonstrate that Talk2Image significantly improves controllability, coherence, and user satisfaction across complex iterative image editing tasks.
\end{itemize}

\section{Related work}
\subsection{Image Generation Models and Ecosystems} Diffusion models \cite{song2020denoising} have become the cornerstone of image generation, enabling high-quality samples through iterative noising and denoising. For text-to-image (T2I) tasks, key frameworks include DALL·E 2 \cite{ramesh2022hierarchical} and Stable Diffusion \cite{rombach2022high}, which use CLIP \cite{radford2021learning} text encoders to condition latent-space generation, and Imagen \cite{saharia2022photorealistic}, which leverages T5 \cite{raffel2020exploring} with cross-attention for textual guidance.
Advances in editing capabilities include ControlNet \cite{zhang2023adding} for spatial condition control, Imagic \cite{kawar2023imagic} for semantic editing via optimization, and DiffEdit \cite{couairon2022diffedit} for automatic region inference. Open-source ecosystems such as HuggingFace, which provides pre-trained models and APIs, and community platforms like Openart and Civitai have accelerated adoption by facilitating workflow sharing. However, these tools still lack robust mechanisms for fine-grained control and multi-turn interaction, limiting their applicability to complex iterative tasks.

\subsection{Dialogue-based Image Generation and Editing}
Dialogue-based image generation and editing systems enable intuitive visual content creation through natural language. Many of these systems use multimodal large language models (MLLMs) to handle both dialogue understanding and image generation/editing tasks. For example, DialogGen~\cite{huang2025dialoggenmultimodalinteractivedialogue} uses a single MLLM, such as Qwen-VL~\cite{bai2023qwenvlversatilevisionlanguagemodel}, for multi-round text-to-image generation, where the MLLM comprehends user input and collaborates with text-to-image models. Similarly, GenArtist~\cite{wang2024genartistmultimodalllmagent} integrates multiple models into a unified framework for image generation and editing, with an MLLM orchestrating task decomposition, tool selection, and execution.
However, these single-agent systems rely on a strictly sequential processing approach, where one MLLM manages both dialogue and image tasks, lacking modularity and effective collaboration.
\subsection{Multi-Agent Systems}
Recent advancements in multi-agent systems (MAS) have enhanced task planning, language collaboration, and multimodal understanding, with frameworks like AutoGPT~\cite{yang2023autogptonlinedecisionmaking} and CAMEL~\cite{li2023camelcommunicativeagentsmind} now incorporating basic support for image generation and visual inputs. Despite these strides, such frameworks remain underoptimized for the unique demands of visual generation and editing tasks. This gap highlights a critical challenge: developing comprehensive multi-agent systems that integrate dynamic labor division, iterative feedback mechanisms, and multimodal perception capabilities specifically tailored to image generation and editing workflows.

\section{Methodology}
\begin{figure*}
    \centerline{\includegraphics[width=\linewidth]{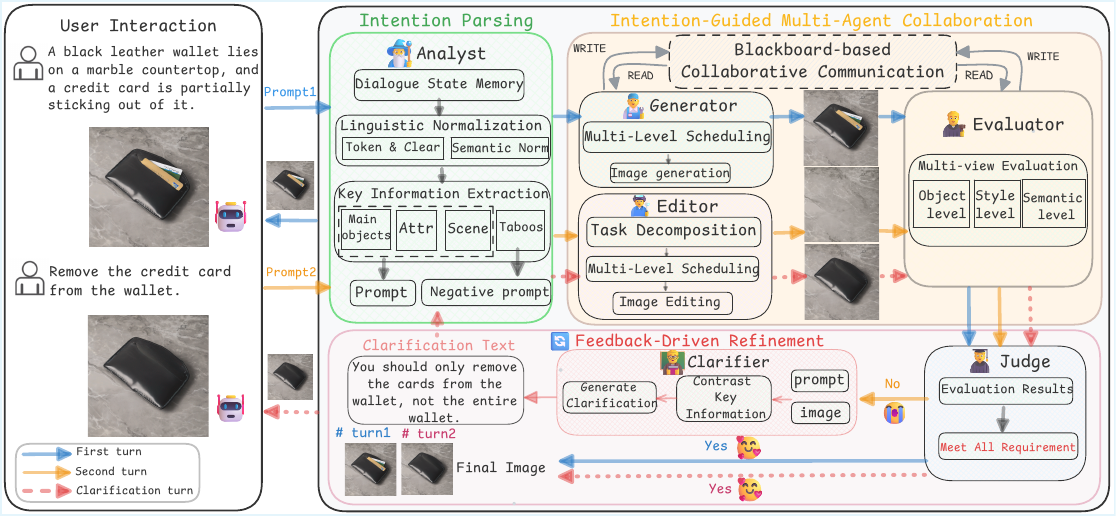}}
    \caption{Overview of Talk2Image, depicting its closed-loop pipeline: intention parsing $\rightarrow$ hierarchical task decomposition $\rightarrow$ multi-agent scheduling \& execution $\rightarrow$ feedback-driven refinement.}
    \label{fig:overview}
\end{figure*}
\textbf{Talk2Image} is a multi-agent system enabling interactive image generation and editing via natural language dialogue. It addresses three core challenges: (1) tracking user intention across turns to mitigate drift; (2) hierarchically decomposing complex tasks into structured subtasks, coordinated by specialized agents to ensure edit coherence; (3) refining outputs through feedback-aware optimization to maintain semantic alignment. The workflow (Figure \ref{fig:overview}) and its formalization in Algorithm \ref{alg:scheduler} implement this hierarchical decomposition, multi-level scheduling, and closed-loop refinement.

\subsection{Dialogue-based Intention Parsing} \label{Intention}
To parse user intention and mitigate intention drift in multi-turn dialogue, Talk2Image employs a four-stage intention parsing pipeline that systematically preserves, normalizes, and structures cumulative user intention. This design addresses historical information loss, linguistic variance, and unstructured representation, which are key drivers of misalignment across turns.

\subsubsection{Dialogue State Memory}
At each turn $t$, the system maintains an evolving interaction history:
\begin{equation}
\mathcal{H}_t = \{(u_1, r_1), (u_2, r_2), \dots, (u_{t-1}, r_{t-1})\}
\label{eq:dialogue-history}
\end{equation}
where $u_i$ and $r_i$ denote the user's instruction and the system's response, respectively. The prompt synthesizer $f_\theta$ fuses the current user input $u_t$ with history to generate a textual state summary:
\begin{equation}
P_t^{\text{text}} = f_\theta(\mathcal{H}_t, u_t)
\label{eq:textual-synthesis}
\end{equation}
The resulting prompt captures the cumulative scene description, resolving user intention revisions such as additions, deletions, or substitutions across turns.

\subsubsection{Linguistic Normalization}
To reduce variance in user phrasing, a canonicalization layer $g(\cdot)$ is applied:
\begin{equation}
P_t^{\text{norm}} = g(P_t^{\text{text}})
\label{eq:normalization}
\end{equation}
This rule-based rewriter eliminates redundancy (e.g., duplicated objects), standardizes phrasing, and clarifies spatial relations, serving as a reliable precursor for structured parsing.

\subsubsection{Key Information Extraction}
The normalized prompt is parsed into structured fields by a semantic extractor $h(\cdot)$:
\begin{equation}
I_t = h(P_t^{\text{norm}}) = (\mathbf{s}, \mathbf{a}, \mathbf{b}, \mathbf{t})
\label{eq:semantic-parse}
\end{equation}
where:
\begin{itemize}
    \item $\mathbf{s} \in \mathcal{S}$: subject entities with optional spatial anchors (e.g., “a cat on the left”).
    \item $\mathbf{a} \in \mathcal{A}$: attribute features (e.g., “white”, “fluffy”).
    \item $\mathbf{b} \in \mathcal{B}$: background scene description (e.g., “grassy field under sunlight”).
    \item $\mathbf{t} \in \mathcal{T}$: disallowed or excluded elements (e.g., “right dog”, “green color”).
\end{itemize}
This intention structure $I_t$ supports downstream modular execution and backward editing.

\subsubsection{Prompt Construction via Template Emission}
A prompt emitter $\Phi(\cdot)$ maps structured intention $I_t$ into three textual directives conforming to a predefined grammar space $\mathcal{G}$:
\begin{equation}
\Phi(I_t) := 
\left\{
\begin{aligned}
\texttt{p}_{\text{pos},t} &= \Phi^+(\mathbf{s}, \mathbf{a}, \mathbf{b}) \in \mathcal{G}_{\text{pos}} \\
\texttt{p}_{\text{neg},t} &= \Phi^-(\mathbf{t}) \in \mathcal{G}_{\text{neg}} \\
\texttt{p}_{\text{scene},t} &= \Phi^{\text{scene}}(\mathbf{b}) \in \mathcal{G}_{\text{scene}}
\end{aligned}
\right.
\label{eq:template-emission}
\end{equation}
Each $\Phi^{*}$ maps symbolic slots into surface realizations using deterministic or learned templates. The grammar domains are defined as:
\begin{align}
\mathcal{G}_{\text{pos}} &::= \texttt{SUBJ}~\texttt{with}~\texttt{ATTR}~\texttt{in}~\texttt{SCENE} \label{eq:grammar-positive} \\
\mathcal{G}_{\text{neg}} &::= \texttt{List}(\texttt{ExcludedItem}) \label{eq:grammar-negative} \\
\mathcal{G}_{\text{scene}} &::= \texttt{SCENE} \quad \text{(no subject placeholders)} \label{eq:grammar-scene}
\end{align}

This emission mechanism ensures consistent formatting, spatial grounding, and decoupled visual control, providing interpretable and editable prompt outputs at each turn.

\begin{algorithm}[t]
\begin{algorithmic}[1]
\Require Dialogue history $\mathcal{H}$, user input $u_t$
\Ensure Final image $\hat{I}_{\text{final}}$
\State $I_t \gets f_\theta(\mathcal{H}, u_t)$ \Comment{Parse into structured intention}
\State $retry \gets 0$
\Repeat
    \State $\mathcal{G}_t \gets \bigcup_{l=1}^L \mathcal{G}_t^{(l)} \gets \Pi(I_t)$ \Comment{Hierarchical goal decomposition}
    \State $G_t = (V_t, E_t) \gets \mathcal{G}(\mathcal{G}_t)$ \Comment{Build DAG of goal dependencies}
    \State $\sigma_t \gets \text{TopologicalSort}(G_t)$ \Comment{Generate valid execution sequence}
    \State $\hat{I}_t \gets \text{Execute}(\sigma_t, \mathcal{B}_t)$ \Comment{Run agents via blackboard scheduling}

    \State Compute $\mathcal{J}_{\text{obj}}, \mathcal{J}_{\text{style}}, \mathcal{J}_{\text{match}}$ \Comment{Multi-view evaluation}
    \State $s \gets \mathcal{J}(\hat{I}_t, I_t)$ \Comment{Aggregate feedback score}
    \If{$s < \tau$}
        \State $I_t \gets A_{\text{clarify}}(\hat{I}_t, I_t)$ \Comment{Refine intention via clarifier}
        \State $retry \gets retry + 1$
    \EndIf
\Until{$s \geq \tau$ or $retry \geq N_{\max}$}
\State \Return $\hat{I}_t$ as $\hat{I}_{\text{final}}$
\end{algorithmic}
\caption{Hierarchical Multi-Agent Execution with Feedback Optimization}
\label{alg:scheduler}
\end{algorithm}

\subsection{Intention-Guided Multi-Agent Collaboration}
Talk2Image employs a heterogeneous multi-agent architecture to enable compositional visual reasoning, integrating three core components: hierarchical task decomposition, multi-level scheduling, and blackboard-based communication. These components collectively support robust multi-turn, feedback-driven visual generation and editing.

\subsubsection{Hierarchical Task Decomposition}
Complex user intentions require decomposition into executable subtasks while preserving coherence. A recursive planner $\Pi: \mathcal{I} \rightarrow 2^{\mathcal{G}}$ breaks down the structured intention $I_t \in \mathcal{I}$ (intention space $\mathcal{I}$) into a hierarchical goal set:

\begin{equation}
\mathcal{G}_t = \bigcup_{l=1}^L \mathcal{G}_t^{(l)}
\end{equation}
Here, $\mathcal{G}_t^{(l)}$ denotes goals at hierarchy level $l$: top-level goals $\mathcal{G}_t^{(1)}$ correspond to primary intention components (e.g., "object replacement"), while lower-level goals $\mathcal{G}_t^{(l)}$ ($l > 1$) represent atomic subtasks (e.g., "locate target") necessary for higher-level goal achievement.

\subsubsection{Multi-Level Scheduling}
Decomposed goals require coordinated execution to satisfy dependencies, optimize agent assignment, and enforce valid order. This is achieved through three interdependent mechanisms:

\textbf{Task-Level Scheduling via DAG.}
Causal dependencies between goals (e.g., "remove before add") are modeled using a directed acyclic graph (DAG) to prevent invalid execution sequences:

\begin{equation}
G_t = (V_t, E_t)
\end{equation}
where $V_t = \{v_g \mid g \in \mathcal{G}_t\}$ (nodes as goals) and $E_t \subseteq V_t \times V_t$ (edges encoding precedence $g \prec h$). The DAG induces a partial order $\prec$ with transitivity 

\begin{equation}
g \prec h \land h \prec k \implies g \prec k
\end{equation}
and irreflexivity ($\neg(g \prec g)$), ensuring feasibility. For $g \in \mathcal{G}_t$, $\text{Pred}(g) = \{h \mid h \prec g\}$ and $\text{Succ}(g) = \{h \mid g \prec h\}$ denote predecessors and successors.

\textbf{Agent-Level Scheduling via Compatibility Matching.}
Each goal $g \in \mathcal{G}_t$ is assigned the optimal agent from pool $\mathcal{A}$ based on capability-requirement alignment. Agents $a \in \mathcal{A}$ have capability vectors $\phi(a) \in \mathbb{R}^d$; goals $g$ have requirement vectors $\psi(g) \in \mathbb{R}^d$. The optimal agent is:

\begin{equation}
a_g = \arg\max_{a \in \mathcal{A}} \texttt{Compat}(\phi(a), \psi(g))
\end{equation}
with the compatibility function implemented via weighted cosine similarity:

\begin{equation}
\texttt{Compat}(x, y) = \frac{x^\top W y}{\|x\|_2 \|y\|_2}
\end{equation}
where $W \in \mathbb{R}^{d \times d}$ (context-adjustable weight matrix) prioritizes relevant capabilities.

\textbf{Execution Scheduling via Topological Sorting.}
To ensure dependencies are met, a topological sort of $G_t$ generates a linear sequence $\sigma_t = [g_1, \dots, g_n]$ where $g_i \prec g_j \implies i < j$. The scheduler $\mathcal{S}$ triggers agents only when predecessors complete:

\begin{equation}
\mathcal{S}(g, G_t, \mathcal{B}_t) = 
\begin{cases} 
a_g(\mathcal{O}_g^*) & \text{if } \mathcal{O}_g^* \subseteq \mathcal{B}_t \\
\perp & \text{otherwise}
\end{cases}
\end{equation}
Here, $\mathcal{O}_g^* = \{o_h \mid h \in \text{Pred}(g)\}$ (required predecessor outputs), $\mathcal{B}_t$ (blackboard), and $\perp$ (pending execution) ensure causal consistency.

\subsubsection{Blackboard-based Collaborative Communication}
Agents share contextual information via a persistent blackboard $\mathcal{B}_t \in \mathcal{B}$ (state space $\mathcal{B}$), a tuple $\langle \mathcal{I}_t^*, \mathcal{O}_t^*, \mathcal{R}_t^* \rangle$ storing intention history, agent outputs, and feedback records. Agents interact via two core operations:
(1) Read: $\texttt{READ}(a, \mathcal{B}_t, \chi)$ (retrieve outputs satisfying predicate $\chi$);
(2) Write: $\texttt{WRITE}(a, \mathcal{B}_t, o)$ (append new outputs to the buffer).


Agent $a$'s output for goal $g$ is:

\begin{equation}
o_g = a\left( \texttt{READ}(a, \mathcal{B}_t, \chi_g) \cup \{g\} \right)
\end{equation}
where $\chi_g(o) = (\exists h \in \text{Pred}(g): o = o_h)$ ensures access to relevant predecessor outputs, enabling loose coupling while preserving dependency integrity.

\begin{table*}[!ht]
\centering
\resizebox{\textwidth}{!}{%
\begin{tabular}{lccccccccc}
\toprule
\multirow{2}{*}{\textbf{Model}} & \multicolumn{4}{c}{\textbf{Task}} & \multicolumn{5}{c}{\textbf{Quantitative Metrics}} \\
\cmidrule(lr){2-5} \cmidrule(lr){6-10}
& \textbf{Generate} & \textbf{Edit} & \textbf{VQA} & \textbf{Chat} 
& \textbf{L2} $\downarrow$ & \textbf{CLIP-I} $\uparrow$ & \textbf{CLIP-T} $\uparrow$ & \textbf{DINO} $\uparrow$ & \textbf{Human} $\uparrow$ \\
\midrule
Null Text Inversion    & $\times$   & \checkmark & $\times$ & $\times$   & 0.0335$_{\uparrow0.0037}$ & 0.8468$_{\downarrow0.0730}$ & 0.2710$_{\downarrow0.0447}$ & 0.7529$_{\downarrow0.1201}$ & 0.7218$_{\downarrow0.1046}$ \\
HIVE                   & $\times$   & \checkmark & \checkmark & $\times$   & 0.0557$_{\uparrow0.0259}$ & 0.8004$_{\downarrow0.1194}$ & 0.2673$_{\downarrow0.0484}$ & 0.6463$_{\downarrow0.2267}$ & 0.6848$_{\downarrow0.1416}$ \\
InstructPix2Pix        & $\times$   & \checkmark & $\times$ & $\times$  & 0.0598$_{\uparrow0.0300}$ & 0.7924$_{\downarrow0.1274}$ & 0.2726$_{\downarrow0.0431}$ & 0.6177$_{\downarrow0.2553}$ & 0.7232$_{\downarrow0.1032}$ \\
MagicBrush             & $\times$   & \checkmark & $\times$ & $\times$   & 0.0353$_{\uparrow0.0055}$ & 0.8924$_{\downarrow0.0274}$ & 0.2754$_{\downarrow0.0403}$ & 0.8273$_{\downarrow0.0457}$ & 0.7864$_{\downarrow0.0400}$ \\
Genartist              & \checkmark & \checkmark & $\times$ & $\times$   & \textbf{0.0298}$_{=}$ & 0.9071$_{\downarrow0.0127}$ & 0.3067$_{\downarrow0.0090}$ & 0.8492$_{\downarrow0.0238}$ & 0.7985$_{\downarrow0.0279}$ \\
\midrule
Talk2Image(Ours)       & \checkmark & \checkmark & \checkmark & \checkmark & \textbf{0.0298} & \textbf{0.9198} & \textbf{0.3157} & \textbf{0.8730} & \textbf{0.8264} \\
\bottomrule
\end{tabular}
}
\caption{Multi-turn editing results. \checkmark\ indicates support for the feature; $\times$ indicates lack of support, and bold indicates the best results. Subscripts show performance differences relative to the best-performing model (arrows: $\uparrow$ worse for lower-better metrics, $\downarrow$ worse for higher-better metrics).}
\label{table:editing}
\end{table*}
\subsection{Multi-View Feedback-Driven Refinement}
To ensure semantic alignment and visual quality, Talk2Image employs a closed-loop optimization mechanism that evaluates generated images from multiple perspectives and refines them incrementally via feedback. This stage builds on parsed intentions and multi-agent execution, using feedback to correct discrepancies between output and user intention.

\subsubsection{Compositive Scoring Mechanism}
Let $\hat{I}_t$ denote the generated image at turn $t$, and $I_t$ the parsed instruction. The multi-view feedback function $\mathcal{J}$ combines three normalized scores ($[0,1]$):
\begin{equation}
\mathcal{J}(\hat{I}_t, I_t) = \lambda_o \cdot \mathcal{J}_{\text{obj}} + \lambda_s \cdot \mathcal{J}_{\text{style}} + \lambda_m \cdot \mathcal{J}_{\text{match}}
\label{eq:evaluation-function}
\end{equation}
where $\mathcal{J}_{\text{obj}}$ is the F1 score between detected objects and target nouns in $I_t$, $\mathcal{J}_{\text{style}}$ denotes cosine similarity with pre-defined reference style embeddings, and $\mathcal{J}_{\text{match}}$ represents CLIP-based alignment between $\hat{I}_t$ and the given textual prompt.

Weights are equal in initialization but adjusted dynamically (e.g., higher $\lambda_s$ for style tasks, $\lambda_o$ for object edits). If $\mathcal{J}(\hat{I}_t, I_t) < \tau$, refinement is triggered.


\subsubsection{Iterative Refinement}
When quality falls below threshold, a Clarifier Agent synthesizes an updated instruction:
\begin{equation}
I_t' = A_{\text{clarify}}(\hat{I}_t, I_t)
\label{eq:clarification-instruction}
\end{equation}
This agent uses image captioning and language modeling to capture discrepancies (e.g., ``Add another apple to the left plate''), enabling a revised task graph to regenerate $\hat{I}_t$. The feedback loop follows:
\begin{equation}
I_t \rightarrow G_t \rightarrow \hat{I}_t \rightarrow \mathcal{J} \rightarrow I_t' \rightarrow \dots
\label{eq:feedback-loop}
\end{equation}

\subsubsection{Adaptive Graph Replanning}
Rather than re-executing the entire pipeline, the system computes intent updates $\Delta I_t$ and fuses them incrementally ($I_t' = I_t \oplus \Delta I_t$), retaining unchanged elements. A new graph $G_t'$ is created, reusing valid subgraphs from $G_t$ to recompute only affected branches, which improves efficiency and convergence.

As shown in Algorithm \ref{alg:scheduler}, the system iteratively refines $I_t$ and $G_t$ until $\hat{I}_t$ meets the quality threshold. The final image is output when the feedback score $s = \mathcal{J}(\hat{I}_t, I_t)$ stabilizes above $\tau$ or the retry limit $N_{\max}$ is reached:
\begin{equation}
\hat{I}_{\text{final}} = \hat{I}_t \big|_{s \geq \tau \lor \text{retry} = N_{\max}}
\end{equation}
This result integrates cumulative intent, coherent multi-agent edits, and feedback-driven refinements, ensuring alignment with conversation history and visual quality.

\subsection{System Perspective}
Talk2Image can be interpreted as a multi-turn interactive framework integrating three core components: LLM-based intention modeling, task-oriented symbolic control, and weakly supervised multi-agent execution. Its modular design and multi-turn interaction mechanism enable seamless integration of diverse AI models, support collaborative human-AI interaction, and facilitate controllable real-time editing. Meanwhile, the feedback-driven cyclic refinement ensures persistent semantic alignment between user inputs and generated images across iterations. Overall, Talk2Image offers a systematic solution for high-quality image generation and fine-grained editing in conversational scenarios.

\section{Experiments}
\subsection{Experimental Setup}

\subsubsection{Datasets.}
We evaluate the single-turn generation capability of Talk2Image on the T2I-CompBench benchmark~\cite{huang2023t2i}. For multi-turn editing, we use the MagicBrush benchmark~\cite{zhang2023magicbrush}, the first large-scale, manually annotated dataset 
specifically designed for multi-turn image editing.


\subsubsection{Baselines.}

We benchmark Talk2Image against three categories of representative baselines to cover both single-turn generation and multi-turn editing settings:

(1) Single-turn Text-to-Image Generation Models, including Stable Diffusion v2~\cite{Rombach_2022_CVPR}, DALL-E 2/3~\cite{ramesh2022hierarchical, openai2023dalle3}, StructureDiffusion~\cite{feng2023trainingfree}, GORS~\cite{huang2023t2i} and SDXL~\cite{podell2023sdxl}. 

(2) Multi-turn Image Editing Models, such as MagicBrush~\cite{zhang2023magicbrush}, InstructPix2Pix~\cite{brooks2023instructpix2pix}, Null Text Inversion~\cite{mokady2023null}, HIVE~\cite{thusoo2009hive} and GenArtist~\cite{wang2024genartist}.

(3) Dialogue-based Multimodal Systems, including SDXL, Seed-X~\cite{ge2024seed}, Hunyuan-DiT ~\cite{li2024hunyuandit} and Qwen2.5-VL~\cite{Qwen2.5-VL}.

\begin{table}[ht]
  \centering
  \resizebox{0.47\textwidth}{!}{%
    \begin{tabular}{lcccc}
    \toprule
    \multirow{2}{*}{\textbf{Model}} & \multicolumn{2}{c}{Spatial relationship} & \multicolumn{2}{c}{Non-Spatial} \\
    \cmidrule(lr){2-3} \cmidrule(lr){4-5}
     & UniDet \(\uparrow\) & Human \(\uparrow\) & CLIP \(\uparrow\) & Human \(\uparrow\) \\
    \midrule
    Stable v2      & 0.1342 & 0.3467 & 0.3127 & 0.5827 \\
    DALL-E 2       & 0.1283 & 0.3640 & 0.3043 & 0.5964 \\
    Structure v2   & 0.1386 & 0.3467 & 0.3111 & 0.6745 \\
    GORS           & 0.1815 & 0.4560 & 0.3193 & 0.6853 \\
    SDXL           & 0.2133 & 0.4425 & 0.3119 & 0.7160 \\
    DALL-E 3       & 0.2543 & 0.4956 & 0.3003 & 0.7255 \\ \hline
    Talk2Image (ours) & \textbf{0.2676} & \textbf{0.5240} & \textbf{0.3269} & \textbf{0.7853} \\
    \bottomrule
    \end{tabular}%
    
  }
\caption{Single-turn generation results on spatial and non-spatial relationships.}
\label{table:gen}
\end{table}

\subsubsection{Implementation Details.}
Talk2Image is a zero-shot framework using Qwen2.5-VL-7B as the agent executor. It runs efficiently on a single NVIDIA A100 (80GB) with no training or fine-tuning required.

\subsubsection{Metrics.}
We use both automatic and human evaluation metrics: (1) For single-turn generation, UniDet accuracy~\cite{zhou2021simple}, CLIP score~\cite{radford2021learning}, and Human score; (2) For multi-turn editing, L2 error, CLIP-I/T, DINO similarity~\cite{zhang2022dino}, and Human score. Both Human scores follow T2I-CompBench (25 prompts/category, 300 images/model) with 3 annotators rating alignment 1–5, then normalized and averaged.



\begin{figure*}[!tb]
    \centering
    \includegraphics[width=\linewidth]{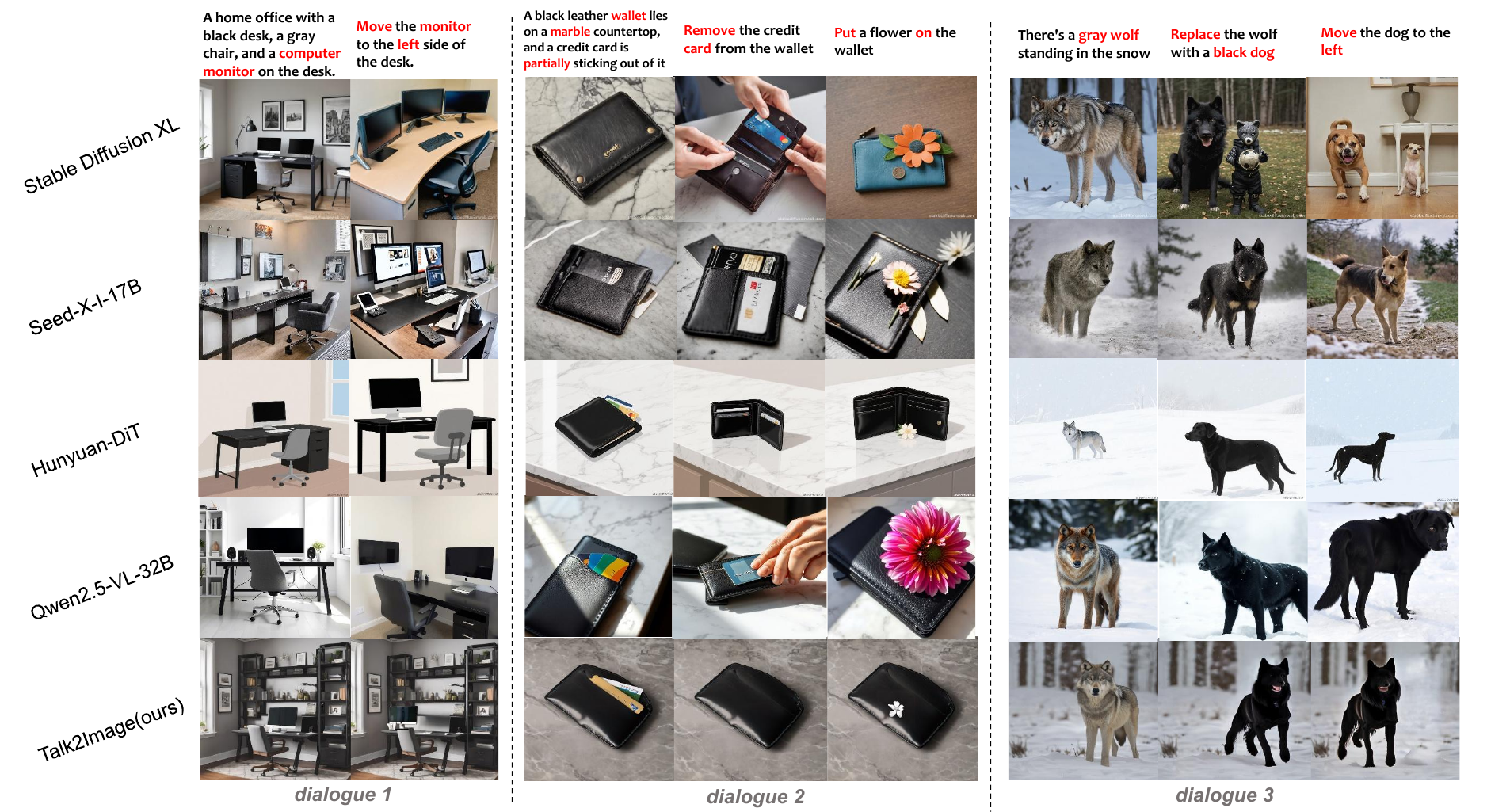}
    \caption{Visualization of Talk2Image Outputs of Stable Diffusion XL, Seed-X-17B, Hunyuan-DiT, Qwen2.5-VL-32B, Talk2Image(ours).}
    \label{fig:quantity}
\end{figure*}
\subsubsection{Supplementary Mechanism and Hyperparameter Validation}
To ensure the robustness of our compositive scoring mechanism and parameter settings, we conduct additional supplementary experiments, detailed in the \textit{Appendix}. 
Specifically, these include: (1) comparative experiments on dynamic weight adjustment strategies, contrasting fixed weights, rule-based dynamic weights, and random dynamic weights; 
(2) sensitivity analysis of key hyperparameters, involving systematic tests on feedback threshold $\tau$ and maximum retry count $N_{\text{max}}$. These experiments further validate the rationality of our design choices.

\subsection{Single-turn Generation Evaluation}  
We evaluate Talk2Image on single-turn text-to-image generation, focusing on spatial relationships and non-spatial attributes. As shown in Table~\ref{table:gen}, it outperforms state-of-the-art T2I models across all metrics: its UniDet score is 5.2\% higher than DALL-E 3 and 25.5\% higher than SDXL in spatial tasks, with human ratings surpassing DALL-E 3 by 5.7\%. For non-spatial attributes, its CLIP score is 2.4\% above GORS, and human ratings are 8.2\% higher than DALL-E 3.  

This superiority stems from two strengths. First, our multi-view feedback mechanism enables comprehensive evaluation (e.g., semantic alignment, visual consistency) and iterative refinement to correct subtle text-image mismatches. Second, integrating specialized models as tools leverages both large foundational models (e.g., DALL-E 3) and task-specific optimizations (e.g., GORS), avoiding single-model limitations. This hybrid framework, combined with self-correction, ensures precise text-image control, which is critical for complex prompts with intricate attribute bindings or spatial relationships.



\subsection{Multi-turn Editing Evaluation}  
We evaluate Talk2Image on multi-turn image editing, where users iteratively refine images via conversational feedback. Table~\ref{table:editing} shows Talk2Image uniquely supports a unified set of capabilities (generation, editing, VQA, chat), while baselines are functionally limited. Quantitatively, it outperforms all baselines, with leads of 1.4\% in CLIP-I, 9.0\% in CLIP-T, and 2.8\% in DINO over Genartist, plus a 3.5\% advantage in human evaluations.  

The core reasons lie in its modular design. The multi-turn intention parser tracks evolving user intention across dialogue turns, effectively mitigating intention drift by maintaining context consistency, ensuring each edit aligns with both current requests and prior interactions. Additionally, multi-agent collaboration enables fine-grained task division: specialized agents handle distinct editing needs (e.g., style adjustments, object additions), while a coordination module integrates their outputs to avoid incoherent edits. Unlike single-model baselines, which struggle with diverse editing demands, Talk2Image leverages the strengths of multiple models through this collaborative framework, adapting flexibly to complex, multi-step refinement tasks.

\subsection{Qualitative Evaluation}

To assess multi-turn coherence and visual controllability, we compare representative outputs from Talk2Image and baselines on a shared set of dialogue prompts.  
As shown in Figure~\ref{fig:quantity}, Talk2Image preserves semantic consistency across turns, accurately accumulates user intention, and maintains visual quality. In contrast, single-agent or one-shot models often suffer from layout drift, or incomplete edits.

For example, in the second turn of the first dialogue (where the monitor is instructed to move to the left side of the desk), baseline models introduce unintended layout changes compared to their initial outputs, while Talk2Image updates only the specified region. Similar inconsistencies are observed in the other two dialogues. This discrepancy stems from the fact that most baselines rely on a single-model pipeline that rewrites prompts and regenerates entire images at each turn, leading to cross-turn misalignment. In contrast, Talk2Image leverages multi-agent collaboration to decompose tasks, precisely localize changes, generate masks, and apply fine-grained edits on the original image, thus ensuring global consistency throughout the dialogue.

\begin{table*}[ht]
\centering
\resizebox{\textwidth}{!}{%
\begin{tabular}{lccccc}
\toprule
\multirow{2}{*}{\textbf{Model}} & \multicolumn{5}{c}{\textbf{Quantitative Metrics}} \\ 
\cmidrule(lr){2-6} 
& \textbf{L2} $\downarrow$ & \textbf{CLIP-I} $\uparrow$ & \textbf{CLIP-T} $\uparrow$ & \textbf{DINO} $\uparrow$ & \textbf{Human} $\uparrow$ \\ 
\midrule
Talk2Image w/o Multi-turn Intention Parser & 0.0310$_{\uparrow0.0012}$ & 0.8988$_{\downarrow0.0210}$ & 0.3022$_{\downarrow0.0135}$ & 0.8542$_{\downarrow0.0188}$ & 0.7968$_{\downarrow0.0296}$ \\
Talk2Image w/o Multi-Agent Collaboration & 0.0315$_{\uparrow0.0017}$ & 0.8962$_{\downarrow0.0236}$ & 0.3015$_{\downarrow0.0142}$ & 0.8356$_{\downarrow0.0374}$ & 0.7862$_{\downarrow0.0402}$ \\
Talk2Image w/o Feedback Refinement     & 0.0308$_{\uparrow0.0010}$ & 0.9011$_{\downarrow0.0187}$ & 0.2948$_{\downarrow0.0209}$ & 0.8412$_{\downarrow0.0318}$ & 0.8031$_{\downarrow0.0233}$ \\
\midrule
\textbf{Talk2Image (Ours)}    & \textbf{0.0298} & \textbf{0.9198} & \textbf{0.3157} & \textbf{0.8730} & \textbf{0.8264} \\
\bottomrule
\end{tabular}
}
\caption{Ablation study on the impact of three core components: Multi-turn Intention Parser, Multi-Agent Collaboration, and Feedback Optimization.}
\label{table:ablation}
\end{table*}

\begin{figure*}[!tb]
    \centering
    \includegraphics[width=\linewidth]{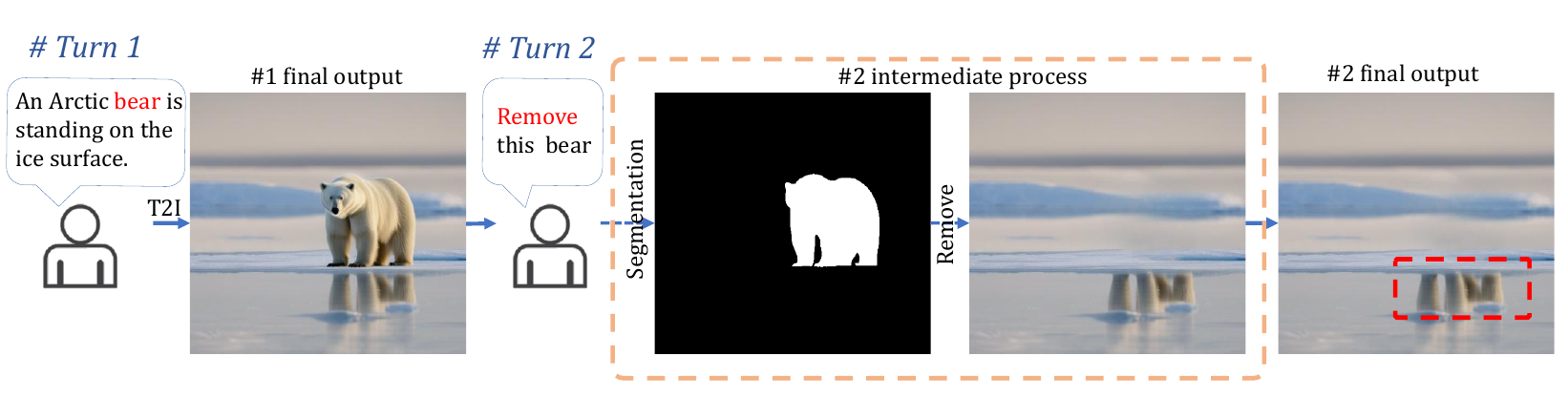}
    \caption{Error Case Example: Incomplete Removal of Bear and Its Reflection.}
    \label{fig:errorcase}
\end{figure*}

\subsection{Ablation Study}
We assess the contribution of three core components via ablation:
the \textit{Multi-Turn Intention Parser}, \textit{Multi-Agent Collaboration} module, and \textit{Multi-View Feedback Refinement} mechanism. Results in Table~\ref{table:ablation} show that removing any component consistently degrades performance across all metrics, confirming their necessity.

\textbf{Multi-Turn Intent Parser.} This component parses user intention by incorporating dialogue history to track evolving goals. Without it, the model suffers from \textit{intention drift} (failing to aggregate instructions over time), leading to a 4.3\% lower CLIP-T score and 2.96-point drop in human ratings \textit{(Line 1 vs. Line 4)}, indicating significant semantic misalignment.

\textbf{Multi-Agent Collaboration.} This module distributes visual subgoals to specialized agents. Removing it results in monolithic generation unable to support localized edits, causing a 4.3\% drop in DINO similarity and 4.9\% lower human preference \textit{(Line 2 vs. Line 4)} due to visual incoherence.

\textbf{Multi-View Feedback Refinement.} This closed-loop mechanism refines images via multi-view evaluation. Without it, the model cannot correct errors, leading to a 2.0\% lower CLIP-I score and 3.4\% higher L2 error \textit{(Line 3 vs. Line 4)}, confirming feedback's role in enhancing alignment and consistency.

\subsection{Limitation and Error Case Analysis}
We examine a representative error case in Talk2Image (Figure~\ref{fig:errorcase}). 
Though the system effectively decomposes intention and schedules tool execution, performance is bounded by underlying models’ capabilities.
For example, in a removal task ("remove the bear"), the segmentation module successfully detects the animal's body but fails to capture its reflection beneath the ice. As a result, the generated image retains a visible shadow even after removal, deviating from the expected outcome.


This case underscores both a limitation and a strength of Talk2Image. It reveals the system’s reliance on third-party tools for fine-grained visual reasoning. However, it also highlights its adaptability: built upon open-source communities, Talk2Image naturally benefits from continual model improvements. 
More importantly, as a no-training framework, Talk2Image allows seamless integration of updated models without retraining, reflecting its modularity and adaptability in rapidly evolving tool communities.

\section{Conclusion}
We present Talk2Image, a multi-agent system for interactive image generation and editing in multi-turn dialogues. By decomposing user intention into structured prompts and leveraging specialized agents for collaborative execution and iterative refinement, it addresses key limitations of prior single-agent approaches.
Experiments confirm its superiority: single-turn generation outperforms DALL-E 3 by 5.2\% (UniDet) and SDXL by 25.5\%; multi-turn editing leads baselines by 1.4-9.0\% in automated metrics, with stronger controllability and consistency. These results validate the potential of modular agent-based systems for intelligent visual dialogue interactions.
Future work will explore heterogeneous topology-based collaboration theories to coordinate agents with diverse functions and roles via heterogeneous edges, enhancing flexibility for dynamic role assignments in complex visual dialogue scenarios.

\bibliography{arxiv}

\begin{thebibliography}{37}
\providecommand{\natexlab}[1]{#1}

\bibitem[{Bai et~al.(2023)Bai, Bai, Yang, Wang, Tan, Wang, Lin, Zhou, and Zhou}]{bai2023qwenvlversatilevisionlanguagemodel}
Bai, J.; Bai, S.; Yang, S.; Wang, S.; Tan, S.; Wang, P.; Lin, J.; Zhou, C.; and Zhou, J. 2023.
\newblock Qwen-VL: A Versatile Vision-Language Model for Understanding, Localization, Text Reading, and Beyond.
\newblock arXiv:2308.12966.

\bibitem[{Bai et~al.(2025)Bai, Chen, Liu, Wang, Ge, Song, Dang, Wang, Wang, Tang, Zhong, Zhu, Yang, Li, Wan, Wang, Ding, Fu, Xu, Ye, Zhang, Xie, Cheng, Zhang, Yang, Xu, and Lin}]{Qwen2.5-VL}
Bai, S.; Chen, K.; Liu, X.; Wang, J.; Ge, W.; Song, S.; Dang, K.; Wang, P.; Wang, S.; Tang, J.; Zhong, H.; Zhu, Y.; Yang, M.; Li, Z.; Wan, J.; Wang, P.; Ding, W.; Fu, Z.; Xu, Y.; Ye, J.; Zhang, X.; Xie, T.; Cheng, Z.; Zhang, H.; Yang, Z.; Xu, H.; and Lin, J. 2025.
\newblock Qwen2.5-VL Technical Report.
\newblock \emph{arXiv preprint arXiv:2502.13923}.

\bibitem[{Brooks, Holynski, and Efros(2023)}]{brooks2023instructpix2pix}
Brooks, T.; Holynski, A.; and Efros, A.~A. 2023.
\newblock Instructpix2pix: Learning to follow image editing instructions.
\newblock In \emph{Proceedings of the IEEE/CVF conference on computer vision and pattern recognition}, 18392--18402.

\bibitem[{Calegari et~al.(2021)Calegari, Ciatto, Mascardi, and Omicini}]{calegari2021logic}
Calegari, R.; Ciatto, G.; Mascardi, V.; and Omicini, A. 2021.
\newblock Logic-based technologies for multi-agent systems: a systematic literature review.
\newblock \emph{Autonomous Agents and Multi-Agent Systems}, 35(1): 1.

\bibitem[{Cardoso and Ferrando(2021)}]{cardoso2021review}
Cardoso, R.~C.; and Ferrando, A. 2021.
\newblock A review of agent-based programming for multi-agent systems.
\newblock \emph{Computers}, 10(2): 16.

\bibitem[{Civitai(2022)}]{civitai2022}
Civitai. 2022.
\newblock Civitai. https://civitai.com/.

\bibitem[{Couairon et~al.(2022)Couairon, Verbeek, Schwenk, and Cord}]{couairon2022diffedit}
Couairon, G.; Verbeek, J.; Schwenk, H.; and Cord, M. 2022.
\newblock Diffedit: Diffusion-based semantic image editing with mask guidance.
\newblock \emph{arXiv preprint arXiv:2210.11427}.

\bibitem[{Face(2018)}]{huggingface2018}
Face, H. 2018.
\newblock Hugging Face. https://huggingface.co.

\bibitem[{Feng et~al.(2023)Feng, He, Fu, Jampani, Akula, Narayana, Basu, Wang, and Wang}]{feng2023trainingfree}
Feng, W.; He, X.; Fu, T.-J.; Jampani, V.; Akula, A.~R.; Narayana, P.; Basu, S.; Wang, X.~E.; and Wang, W.~Y. 2023.
\newblock Training-Free Structured Diffusion Guidance for Compositional Text-to-Image Synthesis.
\newblock In \emph{The Eleventh International Conference on Learning Representations}.

\bibitem[{Ge et~al.(2024)Ge, Zhao, Zhu, Ge, Yi, Song, Li, Ding, and Shan}]{ge2024seed}
Ge, Y.; Zhao, S.; Zhu, J.; Ge, Y.; Yi, K.; Song, L.; Li, C.; Ding, X.; and Shan, Y. 2024.
\newblock Seed-x: Multimodal models with unified multi-granularity comprehension and generation.
\newblock \emph{arXiv preprint arXiv:2404.14396}.

\bibitem[{Ho, Jain, and Abbeel(2020)}]{ho2020denoising}
Ho, J.; Jain, A.; and Abbeel, P. 2020.
\newblock Denoising diffusion probabilistic models.
\newblock \emph{Advances in neural information processing systems}, 33: 6840--6851.

\bibitem[{Huang et~al.(2023)Huang, Sun, Xie, Li, and Liu}]{huang2023t2i}
Huang, K.; Sun, K.; Xie, E.; Li, Z.; and Liu, X. 2023.
\newblock T2i-compbench: A comprehensive benchmark for open-world compositional text-to-image generation.
\newblock \emph{Advances in Neural Information Processing Systems}, 36: 78723--78747.

\bibitem[{Huang et~al.(2024)Huang, Long, Deng, Chu, Xiong, Liang, Cheng, Lu, and Liu}]{huang2024dialoggen}
Huang, M.; Long, Y.; Deng, X.; Chu, R.; Xiong, J.; Liang, X.; Cheng, H.; Lu, Q.; and Liu, W. 2024.
\newblock Dialoggen: Multi-modal interactive dialogue system for multi-turn text-to-image generation.
\newblock \emph{arXiv preprint arXiv:2403.08857}.

\bibitem[{Huang et~al.(2025)Huang, Long, Deng, Chu, Xiong, Liang, Cheng, Lu, and Liu}]{huang2025dialoggenmultimodalinteractivedialogue}
Huang, M.; Long, Y.; Deng, X.; Chu, R.; Xiong, J.; Liang, X.; Cheng, H.; Lu, Q.; and Liu, W. 2025.
\newblock DialogGen: Multi-modal Interactive Dialogue System for Multi-turn Text-to-Image Generation.
\newblock arXiv:2403.08857.

\bibitem[{Kawar et~al.(2023)Kawar, Zada, Lang, Tov, Chang, Dekel, Mosseri, and Irani}]{kawar2023imagic}
Kawar, B.; Zada, S.; Lang, O.; Tov, O.; Chang, H.; Dekel, T.; Mosseri, I.; and Irani, M. 2023.
\newblock Imagic: Text-based real image editing with diffusion models.
\newblock In \emph{Proceedings of the IEEE/CVF conference on computer vision and pattern recognition}, 6007--6017.

\bibitem[{Li et~al.(2023)Li, Hammoud, Itani, Khizbullin, and Ghanem}]{li2023camelcommunicativeagentsmind}
Li, G.; Hammoud, H. A. A.~K.; Itani, H.; Khizbullin, D.; and Ghanem, B. 2023.
\newblock CAMEL: Communicative Agents for "Mind" Exploration of Large Language Model Society.
\newblock arXiv:2303.17760.

\bibitem[{Li et~al.(2024)Li, Zhang, Lin, Xiong, Long, Deng, Zhang, Liu, Huang, Xiao, Chen, He, Li, Li, Zhang, Quan, Lu, Huang, Yuan, Zheng, Li, Zhang, Zhang, Chen, Liu, Fang, Wang, Xue, Tao, Zhu, Liu, Lin, Sun, Li, Wang, Chen, Hu, Xiao, Chen, Liu, Liu, Wang, Yang, Jiang, and Lu}]{li2024hunyuandit}
Li, Z.; Zhang, J.; Lin, Q.; Xiong, J.; Long, Y.; Deng, X.; Zhang, Y.; Liu, X.; Huang, M.; Xiao, Z.; Chen, D.; He, J.; Li, J.; Li, W.; Zhang, C.; Quan, R.; Lu, J.; Huang, J.; Yuan, X.; Zheng, X.; Li, Y.; Zhang, J.; Zhang, C.; Chen, M.; Liu, J.; Fang, Z.; Wang, W.; Xue, J.; Tao, Y.; Zhu, J.; Liu, K.; Lin, S.; Sun, Y.; Li, Y.; Wang, D.; Chen, M.; Hu, Z.; Xiao, X.; Chen, Y.; Liu, Y.; Liu, W.; Wang, D.; Yang, Y.; Jiang, J.; and Lu, Q. 2024.
\newblock Hunyuan-DiT: A Powerful Multi-Resolution Diffusion Transformer with Fine-Grained Chinese Understanding.
\newblock arXiv:2405.08748.

\bibitem[{Ma et~al.(2025)Ma, Zhang, Zhao, Liu, Fan, and Hu}]{ma2025dialogdraw}
Ma, S.; Zhang, X.; Zhao, Z.; Liu, B.; Fan, C.; and Hu, Z. 2025.
\newblock DialogDraw: Image Generation and Editing System Based on Multi-Turn Dialogue.
\newblock In \emph{Proceedings of the AAAI Conference on Artificial Intelligence}, volume~39, 24795--24803.

\bibitem[{Mokady et~al.(2023)Mokady, Hertz, Aberman, Pritch, and Cohen-Or}]{mokady2023null}
Mokady, R.; Hertz, A.; Aberman, K.; Pritch, Y.; and Cohen-Or, D. 2023.
\newblock Null-text inversion for editing real images using guided diffusion models.
\newblock In \emph{Proceedings of the IEEE/CVF conference on computer vision and pattern recognition}, 6038--6047.

\bibitem[{OpenAI(2023)}]{openai2023dalle3}
OpenAI. 2023.
\newblock DALL·E 3 System Card.
\newblock \url{https://openai.com/dall-e-3}.
\newblock Accessed: 2025-07-28.

\bibitem[{OpenArt(2021)}]{OpenArt2021}
OpenArt. 2021.
\newblock OpenArt. https://openart.ai/.

\bibitem[{Podell et~al.(2023)Podell, English, Lacey, Blattmann, Dockhorn, M{\"u}ller, Penna, and Rombach}]{podell2023sdxl}
Podell, D.; English, Z.; Lacey, K.; Blattmann, A.; Dockhorn, T.; M{\"u}ller, J.; Penna, J.; and Rombach, R. 2023.
\newblock Sdxl: Improving latent diffusion models for high-resolution image synthesis.
\newblock \emph{arXiv preprint arXiv:2307.01952}.

\bibitem[{Radford et~al.(2021)Radford, Kim, Hallacy, Ramesh, Goh, Agarwal, Sastry, Askell, Mishkin, Clark et~al.}]{radford2021learning}
Radford, A.; Kim, J.~W.; Hallacy, C.; Ramesh, A.; Goh, G.; Agarwal, S.; Sastry, G.; Askell, A.; Mishkin, P.; Clark, J.; et~al. 2021.
\newblock Learning transferable visual models from natural language supervision.
\newblock In \emph{International conference on machine learning}, 8748--8763. PmLR.

\bibitem[{Raffel et~al.(2020)Raffel, Shazeer, Roberts, Lee, Narang, Matena, Zhou, Li, and Liu}]{raffel2020exploring}
Raffel, C.; Shazeer, N.; Roberts, A.; Lee, K.; Narang, S.; Matena, M.; Zhou, Y.; Li, W.; and Liu, P.~J. 2020.
\newblock Exploring the limits of transfer learning with a unified text-to-text transformer.
\newblock \emph{Journal of machine learning research}, 21(140): 1--67.

\bibitem[{Ramesh et~al.(2022)Ramesh, Dhariwal, Nichol, Chu, and Chen}]{ramesh2022hierarchical}
Ramesh, A.; Dhariwal, P.; Nichol, A.; Chu, C.; and Chen, M. 2022.
\newblock Hierarchical text-conditional image generation with clip latents.
\newblock \emph{arXiv preprint arXiv:2204.06125}, 1(2): 3.

\bibitem[{Rombach et~al.(2022{\natexlab{a}})Rombach, Blattmann, Lorenz, Esser, and Ommer}]{rombach2022high}
Rombach, R.; Blattmann, A.; Lorenz, D.; Esser, P.; and Ommer, B. 2022{\natexlab{a}}.
\newblock High-resolution image synthesis with latent diffusion models.
\newblock In \emph{Proceedings of the IEEE/CVF conference on computer vision and pattern recognition}, 10684--10695.

\bibitem[{Rombach et~al.(2022{\natexlab{b}})Rombach, Blattmann, Lorenz, Esser, and Ommer}]{Rombach_2022_CVPR}
Rombach, R.; Blattmann, A.; Lorenz, D.; Esser, P.; and Ommer, B. 2022{\natexlab{b}}.
\newblock High-Resolution Image Synthesis With Latent Diffusion Models.
\newblock In \emph{Proceedings of the IEEE/CVF Conference on Computer Vision and Pattern Recognition (CVPR)}, 10684--10695.

\bibitem[{Saharia et~al.(2022)Saharia, Chan, Saxena, Li, Whang, Denton, Ghasemipour, Gontijo~Lopes, Karagol~Ayan, Salimans et~al.}]{saharia2022photorealistic}
Saharia, C.; Chan, W.; Saxena, S.; Li, L.; Whang, J.; Denton, E.~L.; Ghasemipour, K.; Gontijo~Lopes, R.; Karagol~Ayan, B.; Salimans, T.; et~al. 2022.
\newblock Photorealistic text-to-image diffusion models with deep language understanding.
\newblock \emph{Advances in neural information processing systems}, 35: 36479--36494.

\bibitem[{Song, Meng, and Ermon(2020)}]{song2020denoising}
Song, J.; Meng, C.; and Ermon, S. 2020.
\newblock Denoising diffusion implicit models.
\newblock \emph{arXiv preprint arXiv:2010.02502}.

\bibitem[{Thusoo et~al.(2009)Thusoo, Sarma, Jain, Shao, Chakka, Anthony, Liu, Wyckoff, and Murthy}]{thusoo2009hive}
Thusoo, A.; Sarma, J.~S.; Jain, N.; Shao, Z.; Chakka, P.; Anthony, S.; Liu, H.; Wyckoff, P.; and Murthy, R. 2009.
\newblock Hive: a warehousing solution over a map-reduce framework.
\newblock 2(2): 1626--1629.

\bibitem[{Wang et~al.(2024{\natexlab{a}})Wang, Li, Li, and Liu}]{wang2024genartist}
Wang, Z.; Li, A.; Li, Z.; and Liu, X. 2024{\natexlab{a}}.
\newblock Genartist: Multimodal llm as an agent for unified image generation and editing.
\newblock \emph{Advances in Neural Information Processing Systems}, 37: 128374--128395.

\bibitem[{Wang et~al.(2024{\natexlab{b}})Wang, Li, Li, and Liu}]{wang2024genartistmultimodalllmagent}
Wang, Z.; Li, A.; Li, Z.; and Liu, X. 2024{\natexlab{b}}.
\newblock GenArtist: Multimodal LLM as an Agent for Unified Image Generation and Editing.
\newblock arXiv:2407.05600.

\bibitem[{Yang, Yue, and He(2023)}]{yang2023autogptonlinedecisionmaking}
Yang, H.; Yue, S.; and He, Y. 2023.
\newblock Auto-GPT for Online Decision Making: Benchmarks and Additional Opinions.
\newblock arXiv:2306.02224.

\bibitem[{Zhang et~al.(2022)Zhang, Li, Liu, Zhang, Su, Zhu, Ni, and Shum}]{zhang2022dino}
Zhang, H.; Li, F.; Liu, S.; Zhang, L.; Su, H.; Zhu, J.; Ni, L.~M.; and Shum, H.-Y. 2022.
\newblock DINO: DETR with Improved DeNoising Anchor Boxes for End-to-End Object Detection.
\newblock arXiv:2203.03605.

\bibitem[{Zhang et~al.(2023)Zhang, Mo, Chen, Sun, and Su}]{zhang2023magicbrush}
Zhang, K.; Mo, L.; Chen, W.; Sun, H.; and Su, Y. 2023.
\newblock Magicbrush: A manually annotated dataset for instruction-guided image editing.
\newblock \emph{Advances in Neural Information Processing Systems}, 36: 31428--31449.

\bibitem[{Zhang, Rao, and Agrawala(2023)}]{zhang2023adding}
Zhang, L.; Rao, A.; and Agrawala, M. 2023.
\newblock Adding conditional control to text-to-image diffusion models.
\newblock In \emph{Proceedings of the IEEE/CVF international conference on computer vision}, 3836--3847.

\bibitem[{Zhou, Koltun, and Kr{\"a}henb{\"u}hl(2022)}]{zhou2021simple}
Zhou, X.; Koltun, V.; and Kr{\"a}henb{\"u}hl, P. 2022.
\newblock Simple multi-dataset detection.
\newblock In \emph{CVPR}.

\end{thebibliography}

\end{document}